\documentclass[runningheads]{llncs}
\usepackage[T1]{fontenc}

\usepackage{amsmath,amsfonts,bm}

\def\eqref#1{equation~\ref{#1}}

\def\1{\bm{1}}

\DeclareMathAlphabet{\mathsfit}{\encodingdefault}{\sfdefault}{m}{sl}
\SetMathAlphabet{\mathsfit}{bold}{\encodingdefault}{\sfdefault}{bx}{n}

\usepackage{makecell}
\usepackage[utf8]{inputenc} 
\usepackage[T1]{fontenc}    
\usepackage{url}            
\usepackage{booktabs}       
\usepackage{amsfonts}       
\usepackage{nicefrac}       
\usepackage{microtype}      
\usepackage{xcolor}         
\usepackage{xspace}       
\usepackage{multirow}       
\usepackage{graphicx}
\usepackage[x11names]{xcolor}   
\usepackage{bold-extra}
\usepackage{subcaption}
\usepackage{amsmath}

\usepackage{xcolor} 
\usepackage[
    colorlinks=true,
    linkcolor=black,       
    citecolor=black,      
    filecolor=magenta,    
    urlcolor=blue         
]{hyperref}

\definecolor{softred}{HTML}{E04B4B}   
\definecolor{softblue}{HTML}{5068D8}  

\newcommand{\eg}{\emph{e.g.,}\ }

\newcommand{\sect}[1]{\S\ref{#1}}
\newcommand{\sectpar}[1]{(\S\ref{#1})}

\usepackage{inconsolata}   

\newcommand{\all}{\texttt{trace}\xspace}
\newcommand{\alldownsampled}{\texttt{trace downsampled}\xspace}
\newcommand{\last}{\texttt{last}\xspace}
\newcommand{\synthetic}{\texttt{synthetic}\xspace}
\newcommand{\syntheticdownsampled}{\texttt{synthetic downsampled}\xspace}
\newcommand{\pc}{\textsc{Pencil Code}\xspace}

\newcommand{\seenstudentseenprogram}{\texttt{seen student/seen title}\xspace}
\newcommand{\seenstudentunseenprogram}{\texttt{seen student/unseen title}\xspace}
\newcommand{\unseenstudentseenprogram}{\texttt{unseen student/seen title}\xspace}
\newcommand{\unseenstudentunseenprogram}{\texttt{unseen student/unseen title}\xspace}
\newcommand{\gpt}{\texttt{GPT-2}\xspace}
\newcommand{\olmo}{\texttt{OLMo-2}\xspace}

\newcommand{\programtitle}[1]{\texttt{#1}}
\newcommand{\tracetitle}[1]{\texttt{#1}}

\usepackage{hyperref}
\usepackage{url}
\usepackage{caption}
\title{Modeling Student Learning with \\ 3.8 Million Program Traces}

\author{
Alexis Ross\thanks{Equal contribution}\inst{1}\orcidID{0009-0005-1861-0928} \and
Megha Srivastava*\inst{2}\orcidID{0000-0003-0985-7062} \and
Jeremiah Blanchard\inst{3}\orcidID{0000-0003-2995-5102} \and 
Jacob Andreas\inst{1}\orcidID{0000-0002-3141-5845}
}

\authorrunning{A. Ross and M. Srivastava et al.}

\institute{
MIT CSAIL \\
\email{alexisro@mit.edu, jda@mit.edu}
\and
Stanford University \\
\email{megha@cs.stanford.edu}
\and
University of Florida \\
\email{jjb@eng.ufl.edu}
}

\begin{document}

\maketitle

\begin{abstract}
    As programmers write code, they often edit and retry multiple times, creating rich ``interaction traces'' that reveal how they approach coding tasks and provide clues about their level of skill development. For novice programmers in particular, these traces reflect the diverse reasoning processes they employ to code, such as exploratory behavior to understand how a programming concept works, re-strategizing in response to bugs, and personalizing stylistic choices.
    In this work, we explore what can be learned from training language models on such reasoning traces: not just about code, but about coders, and particularly students learning to program.
    We introduce a dataset of over 3.8 million programming reasoning traces from users of \href{https://pencilcode.net/}{\pc}, a free online educational platform used by students  to learn simple programming concepts. Compared to models trained only on final programs or synthetically-generated traces, we find that models trained on real traces are stronger at modeling diverse student behavior. Through both behavioral and probing analyses, we also find that many properties of code traces, such as goal backtracking or number of comments, can be predicted from learned representations of the students who write them. Building on this result, we 
    demonstrate potential to help students recover from mistakes
    by steering code generation models to identify a sequence of edits that will 
    result in more correct code while remaining close to the original student's style.
    Together, our results suggest that many properties of code are properties of individual students and that training on edit traces can lead to models that are more steerable, more predictive of student behavior while programming, and better at generating programs in their final states\footnote{Code and data is available at \url{https://github.com/meghabyte/pencilcode-public}.}.
    \end{abstract}


\begin{figure*}[ht!]
    \centering
    \includegraphics[width=0.99\linewidth]{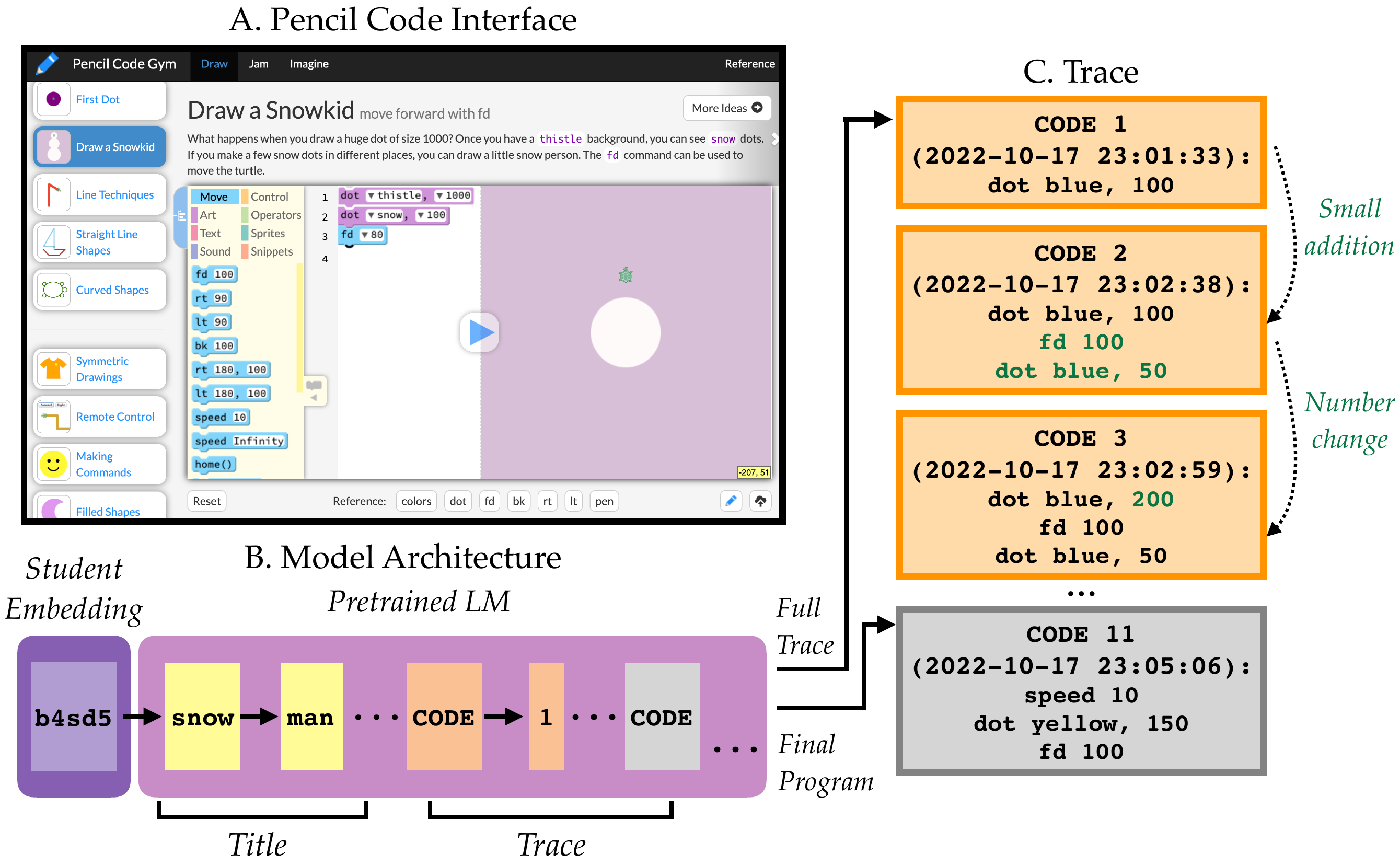}
    \caption{(A) User interface of \pc, where users can program with visual block coding. 
    (B) Our model architecture, with an embedding layer for student IDs. 
    (C) An example trace written for \programtitle{snowman}, along with edit types.}
    \label{fig:overview}
    \vspace{-0.3cm} 
\end{figure*}

\section{Introduction}
Imagine a student learning to code through a simple visual assignment, such as drawing a snowman (Figure~\ref{fig:overview}). They might begin by experimenting with how to draw a single circle, then attempt to assemble the full figure. After running the program and encountering unexpected output, the student may realize gaps in their understanding (e.g., 2D coordinates) and seek help. Once the core logic is resolved, they may personalize the program with comments or colors. Observing only the final code reveals little about this iterative reasoning process.

Although users employ diverse strategies while problem-solving, few public datasets capture this underlying process. This creates a gap in the dominant paradigm of internet-scale pretraining: current data models what users produce, but not \emph{how} they produce it. As a result, state-of-the-art models often adopt non-human-like problem-solving strategies \cite{embers}. While recent work has explored training models to generate ``reasoning'' traces \cite{cot,star},
these efforts primarily aim to improve performance rather than to accurately model human reasoning.

At the same time, a paradigm shift is underway in how users interact with AI assistants, particularly for code generation. Tools such as Cursor and GitHub Copilot integrate directly into development environments, enabling the capture of rich interaction data that records entire coding processes, including attempts, revisions, and assistant interactions. Such data raises new questions: what can models learn from full development traces, and can they capture how different individuals approach the same task? This is especially relevant in educational settings, where iterative debugging and refactoring are central to learning \cite{berland2013using}.

In this work, we curate a dataset of 3.8M \textit{program traces} from real students learning to code on \pc, spanning 9 years and over 1M unique students. The dataset covers a wide range of assignments, from simple graphical tasks like \texttt{snowman} to more complex algorithms such as \texttt{search}. Each trace contains a student ID, assignment title, and a temporally ordered sequence of executed program states. Unlike prior datasets that focus on IDE-level edits \cite{brown2018blackbox} or final submissions \cite{eliseeva2023falconcode}, our dataset is, to our knowledge, the first large-scale collection of execution-bounded code edit sequences suitable for language model training.

We compare three pretraining approaches: training on real program traces (\all), on synthetically generated traces derived from final programs (\synthetic), and on final program states only (\last). Using both \textbf{\emph{behavioral}} and \textbf{\emph{representational}} evaluations, we study how well these models capture both students' final programs and the processes that produced them. Our results show that training on real traces yields richer models of both programs \sectpar{s:general_behavior} and student behaviors \sectpar{s:student_embeddings}. Learned student representations encode meaningful information about reasoning-related properties (\eg deviation frequency, time spent), as well as stylistic features (\eg comments; \sect{s:code_embeddings}, \sect{s:student_embeddings}). Many of these properties can be inferred efficiently from limited student data \sectpar{s:finetuning}, and we demonstrate how they can be leveraged to help students recover from errors \sectpar{s:error_recovery}.

\section{PencilCode Overview}
\label{s:pencilcode}



\href{pencilcode.net}{\pc}, introduced by \cite{pencilcode}, is an open-source educational platform for programming that supports creative projects spanning turtle graphics, music composition, speech synthesis, networking, and interactive storytelling. It utilizes Droplet, a dual-modality code editor that allows users to write code through either a visual block-based interface (similar to Scratch) or directly in web programming languages like CoffeeScript, JavaScript, HTML, and CSS. The platform's block-and-text interface has been shown to support novice programmers' development of expertise \cite{blanchard-amphibian-cs1} and to enhance students' programming skills and attitudes \cite{Deng2020pencilcode}.  We construct a dataset of 3.8M programming traces written on \pc from 2015 to 2024. Each trace consists of a hashed student ID, title (\eg \texttt{snowman}), and an ordered of sequence of programs written by the student, along with associated timestamps. The overall dataset has size 248GB, consisting of  1.3M unique usernames and 3.8M unique (username, program\_name) pairs. The dataset contains an average of 2.86 program traces per user.


\section{Experimental Set-Up}

\paragraph{Models} 
We train 5 LMs on \pc data with continued pretraining. Our main experiments are conducted with a base 124M parameter \gpt model \cite{gpt2}.
However, we also run experiments with a 1B parameter \olmo model \cite{olmo20242olmo2furious}, for which we observe similar results, reported in Table~\ref{tab:olmo_results}. 
The \textbf{\all} model is trained on full traces, while the \textbf{\last} model is trained on only the last program in each trace. Similar to \cite{piterbarg2025synthetic}, we also train a \textbf{\synthetic} model that is trained on traces that are synthetically generated based on the last program in each trace.
In addition, we train \textbf{\alldownsampled} and \textbf{\syntheticdownsampled} models on versions of the \all and \synthetic datasets that are downsampled to match the same number of unique tokens as in the \last dataset. 
We train models with a \textbf{student embedding} layer that maps a student ID to a 768-dimension embedding, which is introduced as a ``soft token'' at the start of all program sequences (Figure \ref{fig:overview}). Introducing an explicit student token enables analysis of behavior at the student level \sectpar{s:student_embeddings}, as well as light-weight adaptation to new students \sectpar{s:finetuning}. 





\begin{figure}[t!]
    \centering
    \includegraphics[width=0.99\linewidth]{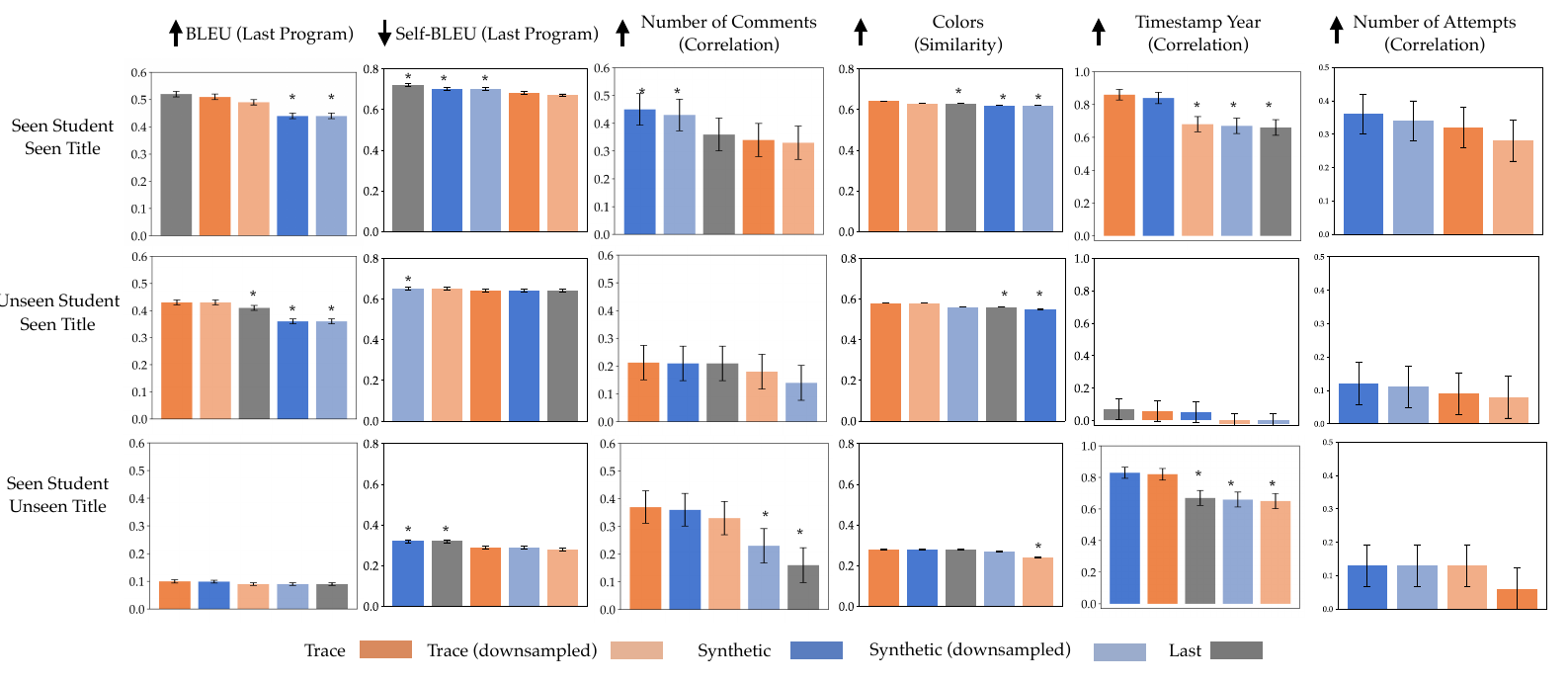}
    \caption{\textbf{Correlation of Generated Trace Properties with Ground Truth (Final Program State)} We evaluate the generated final program state of a trace from sampling all models across evaluation splits. Correlation denotes Pearson's coefficient. The Colors metric compares the cosine similarity of between program color embeddings. * indicates a statistically significant difference with the \all model using a paired T-test between unique (student, title) pairs at $p=0.05$ with Bonferroni correction, and error bars indicate standard errors of the mean.}
    \label{fig:behavior-last}
\end{figure}

\paragraph{Evaluation Splits} 
We study various kinds of generalization of the above models \sectpar{s:generalization}: generalization in-distribution to new (student, title) pairs (where each has been seen before separately), as well as out-of-distribution generalization to unseen students and titles. We create 4 test sets reflecting each kind of generalization. We first hold out 2\% of student IDs and 2\% of trace titles. We then take 80\% of the non-held-out data 
and designate them as training data, then create the in-distribution test set by taking all remaining non-held-out data.
We refer to this split as \seenstudentseenprogram. We then create 3 additional test sets that differ in whether the student or title is held out: \seenstudentunseenprogram 
 (71,799 traces),
\unseenstudentseenprogram (259,345 traces), and
\unseenstudentunseenprogram (8,814 traces). 

\subsection{Evaluation Methods}
\label{s:evaluation_methods}
We consider two types of evaluations of our 5 models: The first is a \textbf{behavioral} evaluation, where we generate Monte Carlo samples with a model and analyze properties of the generated programs (\sect{s:general_behavior}, \sect{s:finetuning}, \sect{s:error_recovery}).\footnote{For all behavioral evaluations, we generate using nucleus sampling with $p=0.9$. 
}  
The second evaluation type is \textbf{representational}, where we probe learned code and student embeddings to understand what information they encode (\sect{s:code_embeddings}, \sect{s:student_embeddings}).
For both, we analyze a variety of properties of code written by students. 

\paragraph{Properties of Programs}
\label{s:program_properties}
For each program that is part of a trace, we measure \textbf{successful execution} (whether a program executes without errors), the \textbf{time} the program was executed on the \pc server, and the number of occurrences of certain \textbf{keywords} (\eg the word \texttt{turtle}, which is associated with \emph{turtle graphics} programs, words associated with \emph{colors} such as \texttt{magenta}), and \emph{comments} (\eg lines starting with \#). We measure these properties for each program in a trace, although we are particularly interested in the last program (which we take as the student's goal state).

\paragraph{Properties of Traces}
\label{s:trace_properties}
We also analyze properties of {traces}, which characterize the full sequence of program code and edits. For the program metrics above, we can consider the mean value across all programs in a trace. In addition, we can look at trace-specific properties including \textbf{goal backtracking}:  We measure the \emph{goal backtracking ratio} of a trace, which is the average fraction of times in a trace that a student's edit results in an \textit{increase} in edit distance between the current program and goal program state. We also measure the counts of \textbf{edit types} in a trace: {small/large additions}, {small/large deletions}, {color/number changes}, and {comment/function additions}.

\paragraph{Behavioral Evaluation Metrics}
While for our representational evaluations (\sect{s:code_embeddings}, \sect{s:student_embeddings}) we can directly measure a probe model's ability to predict the properties of code listed above, for our behavioral evaluations, we \emph{compare} generated samples against ground truth traces written by students. We report the Pearson \textbf{correlations} between these values for each metric. We additionally measure the \textbf{BLEU} score \cite{bleu} to directly compare the similarity of generated program traces against the ground truth trace for a given student ID and title, as well as the \textbf{Self-BLEU} \cite{selfbleu} across the final programs of repeated generated samples. Whereas BLEU captures how close a program is to a reference, Self-BLEU measures how similar a set of generated samples is, with a lower value indicating higher diversity. For both metrics, we average across ${1, 2, 3, 4}-$ngram scores.


\begin{figure}[t!]
    \centering
    \includegraphics[width=0.99\linewidth]{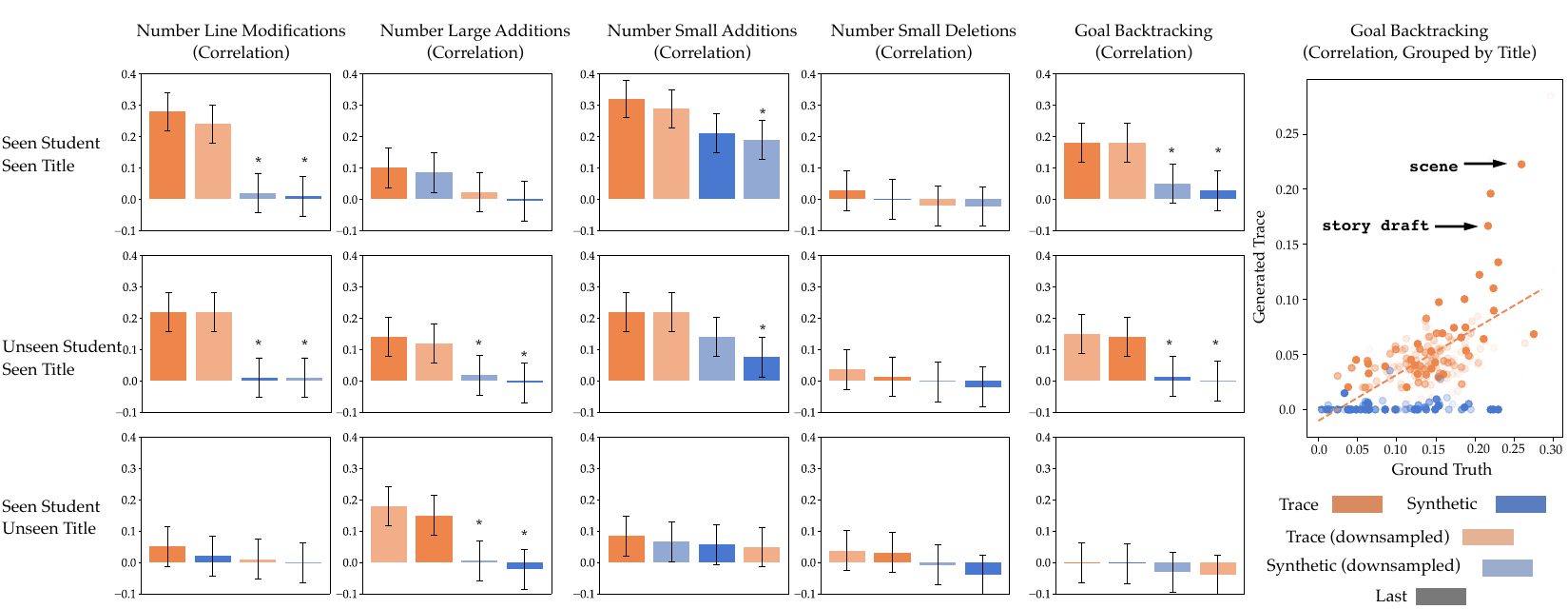}
    \caption{\textbf{Correlation of Generated Trace Properties with Ground Truth (Full Program Trace)} We evaluate generated program traces by sampling all models except \last across evaluation splits. Correlation is measured using Pearson’s coefficient. * denotes a statistically significant difference from the \all model using a paired t-test over (student, title) pairs at $p=0.05$ with Bonferroni correction. Error bars show standard errors of the mean. The right scatter plot illustrates how the \all model captures student goal backtracking, with higher correlations for titles more common in the training data (higher opacity).}
    \label{fig:behavior-edit}
\end{figure}

\section{Experiments}
First, we investigate whether models trained on real edit traces learn richer representations than those trained on synthetic traces or final programs (Sections 4.1-4.3). Second, we examine whether these representations encode student-specific information that enables efficient personalization (Section 4.4). Finally, we demonstrate how learned representations can be applied to practical educational scenarios like style-preserving error recovery (Section 4.5).

\subsection{Modeling General Behavior}
\label{s:general_behavior}

We first ask whether code generated by models is reflective of the programming behaviors of \pc students. We select 200 titles that correspond to assignments found in online resources from \pc.\footnote{We derive them from external resources on learning to code in \pc, including an associated primer (\url{https://book.pencilcode.net/}). 100 of these were seen during training, and 100 were not. } 
For each title, we randomly select 50 students, split evenly between the seen and unseen student splits. We then use the student ID and title to construct a prefix which we use to conditionally generate $n=20$ random samples from each model. We then analyze each sampled trace (or single program in the case of the \last model) for properties described in Section \ref{s:evaluation_methods}.

\begin{table}[h!]
\caption{BLEU score evaluation results for the 1B \olmo model on different splits. The best-performing cell for each column is bolded. We evaluate the best-performing checkpoints for each model after up to 3 epochs of training. Results are for 100 randomly sampled titles corresponding to \pc{} assignments. We sample 10 generations for each title and model.} 

\centering
\footnotesize
\begin{tabular}{lccc}
\hline
\textbf{Model} & \makecell{\texttt{seen student}\\\texttt{seen title}} & \makecell{\texttt{unseen student}\\\texttt{seen title}} & \makecell{\texttt{seen student}\\\texttt{unseen title}} \\
\hline
{\last} & 0.262 $\pm$ 0.013 & 0.246 $\pm$ 0.013 & 0.042 $\pm$ 0.005 \\
{\alldownsampled} & \textbf{0.284} $\pm$ 0.013 & \textbf{0.276} $\pm$ 0.013 & \textbf{0.059} $\pm$ 0.006 \\
\hline
\end{tabular}
\label{tab:olmo_results}
\end{table}

While we present results for a \gpt model in this section, Table~\ref{tab:olmo_results} shows that we see similar generalization trends for a 1B \olmo model. Across all splits, the \alldownsampled model generates final program states that are more similar to ground truth generated programs than those generated by the \last model, suggesting that the gap between trace and \last model increases with scale. 

\paragraph{Generalization In-Distribution} As shown in Figure \ref{fig:behavior-last} (first row), for students and titles that were seen separately during training (but not together), the \all model generates final programs that are more similar (BLEU
) to the ground truth than the \synthetic model and comparable to the \last model. However, the \all model results in higher diversity between generated last programs (lower self-BLEU score) than both the \last and \synthetic models. These results hold even when we control for the number of tokens (\alldownsampled and \syntheticdownsampled), suggesting that training on traces can leader to stronger and richer final program states.

\paragraph{Generalization Out-of-Distribution} 
\label{s:generalization}
The second and third rows in Figure~\ref{fig:behavior-last} present results analyzing the generated final programs for splits where either the student or trace title was unseen. These results reveal how models leverage student IDs and titles: for example, knowing the student ID leads models to generate programs with the correct timestamp years, while knowing the title (\eg \texttt{rose}) is sufficient for using the appropriate colors in graphical programs.
We observe that BLEU significantly drops across all models for the \seenstudentunseenprogram split; thus, despite models' natural language pretraining (prior to training on \pc), generalization is still difficult. However, we observed that titles in this split that \emph{do} reflect program semantics (\eg \texttt{spike function}) result in uniformally high BLEU scores, suggesting some degree of OOD generalization. 

\paragraph{Student Edit Behavior} 
\label{s:edits}
Finally, we ask if generated program traces reflect students' \textit{edit} behavior. Figure \ref{fig:behavior-edit} shows that goal backtracking in generated traces from the \all model are indeed correlated with ground truth metrics, but this correlation slightly decreases when the student is unseen. This suggests that title and student ID \textit{both} play a role in determining whether there is backtracking from the final program in the trace. We also see in the rightmost scatter plot that the correlation of degree of backtracking between generated and ground truth traces
is higher for titles that are more frequent in the training data (\eg \tracetitle{scene}). 
As expected, the \synthetic model only shows high correlation for the ``small addition'' types of edits, which are the only kind it sees during training. 


\begin{figure*}[t]
    \centering
    \includegraphics[width=\columnwidth]{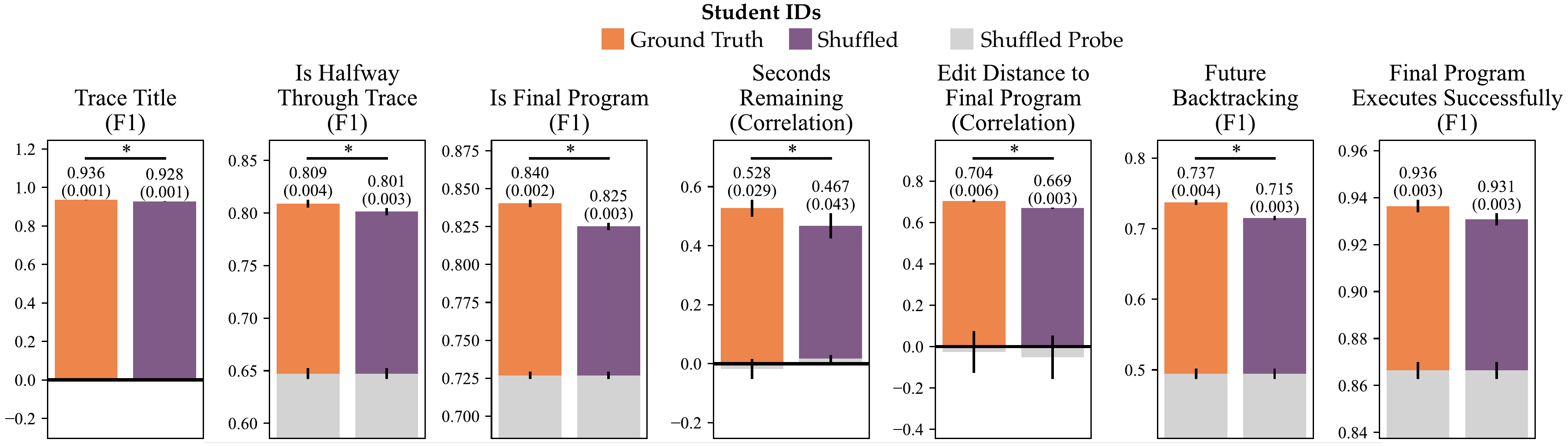}
    \caption{\textbf{Probing Code Representations Given Student IDs and Trace Prefixes.} We report mean F1 scores / Pearson correlations for probes trained on 5 random data splits. Error bars indicate standard errors of the mean. \emph{Shuffled Probe} corresponds to the control where we shuffle inputs/outputs to the probe. * indicates statistically significant differences between the ground truth student and shuffled student probes under a paired T-test between random probes.}
    \label{fig:probing_code_embeddings}
\end{figure*}


\begin{figure*}[t]
    \centering
    \includegraphics[width=\columnwidth]{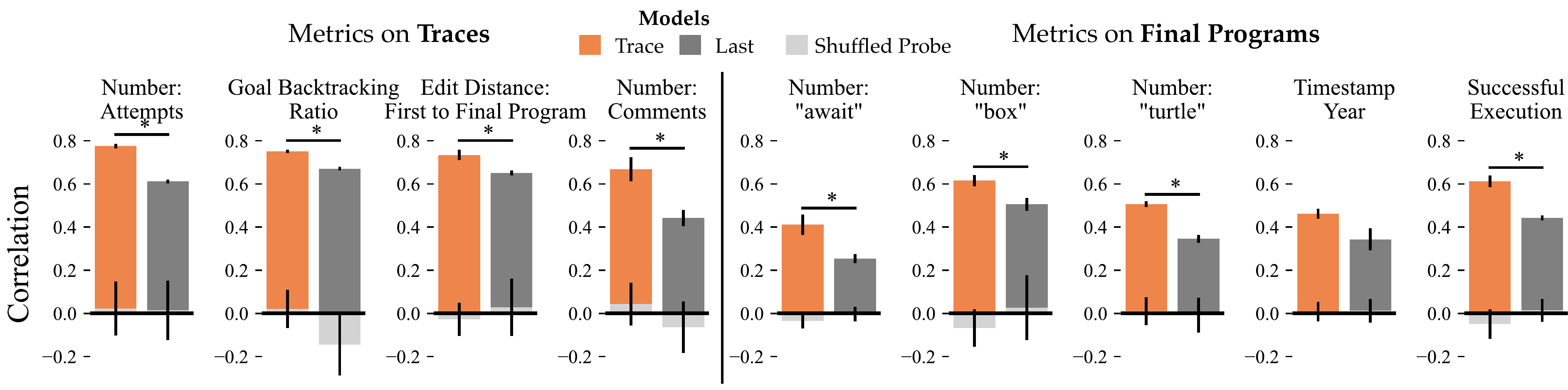}
    \caption{\textbf{Probing Student Representations to Predict Means Across Traces for a Student.} We train probes to predict, for a given metric and student, the mean value of the metric across all of the student's traces. We report mean Pearson correlations for probes trained on 5 random data splits. Error bars indicate standard errors of the mean. \emph{Shuffled Probe} corresponds to the control where we shuffle inputs/outputs to the probe.  * indicates statistically significant differences between the ground truth student and shuffled student probes under a paired T-test between random probes. See \sect{s:student_embeddings} for details.
    }
    \label{fig:probing}
\end{figure*}


\subsection{Probing Code Representations}
\label{s:code_embeddings}

\sect{s:general_behavior} presents evidence that the \all model can predict aspects of students' behavior in its generations.
In this section, we ask whether such information is directly encoded in the model's representations of code: Given a code snippet written by a student, can a probe trained on the \all model's representations predict \emph{future} student behavior, \eg whether the student will backtrack from their goal in the future? To what extent is the performance of the probe (and therefore the information in code representations) dependent on the model's learned information about the particular student?

We construct our probing dataset by first considering the same restricted set of 100 \pc assignment titles that were seen during training (as described in \sect{s:general_behavior}). For each title, we randomly sample 50 traces from the \seenstudentseenprogram split. Additionally, we only analyze students with between 20 and 200 traces in the training dataset for the \all model to ensure that sufficient traces were seen to learn meaningful representations.\footnote{A few IDs had more than 200 traces and corresponded to classrooms with multiple students using the same account, which we deduced from program comments.}
We then construct a probing dataset where inputs are embeddings of traces consisting of varying numbers of programs, and outputs are various code properties.  We train ridge regression/classification probes to predict the \emph{trace title}, whether the program \emph{is the last program}, and whether the program is at least \emph{halfway through the trace}. We also predict more fine-grained future behavior from the student, including whether the student will \emph{backtrack} from the goal/final program later in the trace, the number of \emph{future attempts} the student will make, the number of \emph{seconds} the student will continue to spend, the \emph{edit distance} between the current program state and the final program state, 
and the eventual correctness of the final program in the trace. 

To evaluate the impact of student IDs, we construct input embeddings from shuffled student IDs.
We also shuffle the embeddings 
themselves as a control task for how well probes perform when not using the actual representations \cite{hewitt-liang-2019-designing}.

\paragraph{Results} Figure~\ref{fig:probing_code_embeddings} shows mean F1 scores / Pearson correlations for probes trained across 5 random train/test splits. We find that for all metrics, probes significantly outperform the shuffled probe controls. This suggests that the embeddings contain nontrivial information about code and future student behavior. We also find that for most metrics, conditioning on the true student ID leads to statistically significant better performance than conditioning on shuffled student IDs (except for successful execution of the final program). 
Our results suggest that the \all model not only leverages information about individual students in reasoning about future code steps, but also its embeddings might support educational feedback, \eg intervening when a student is about to backtrack from their goal.

\subsection{Probing Student Representations}
\label{s:student_embeddings}

\sect{s:general_behavior} and \sect{s:code_embeddings} show both behaviorally and representationally that the \all model uses information about student IDs to make predictions about their behavior.
In this section, we directly probe student representations to evaluate what they capture about a student's programming style and abilities. 

We construct the student probing dataset by randomly sampling 2,000 students, each with between 20 and 200 traces. For each student and each property of interest, we compute a ground-truth student-specific target by averaging the metric across all traces written by that student. We then obtain a student embedding by extracting the learned embedding vector for the student from the \all or \last model, and pair this embedding with the corresponding target value. We train MLP probes on 5 random train/test splits. As in \sect{s:student_embeddings}, we also train probes on control datasets in which student embeddings are randomly shuffled, providing a baseline that reflects probe performance when relying only on target metric statistics rather than information encoded in the embeddings \cite{hewitt-liang-2019-designing}.

\paragraph{Results} As shown in Figure \ref{fig:probing}, for both the \all and \last model, probes trained on ground truth student representations outperform the probes trained on shuffled student representations. This suggests that the student representations in both models encode nontrivial information about students. However, the \all model's student representations consistently lead to statistically significant better performance than the \last model's student representations (except for timestamp year), highlighting that the \all model has learned about student behavior beyond what can be learned just from the kinds of programs students write. Even for metrics directly related to the kinds of programs students write (\eg occurrences of \texttt{await}), which could be learned by the \last model, the \all model's student representations lead to better performance, suggesting that by training on traces, the \all model has learned richer information about students.

\begin{figure}[h]
    \centering
    \includegraphics[width=\textwidth]{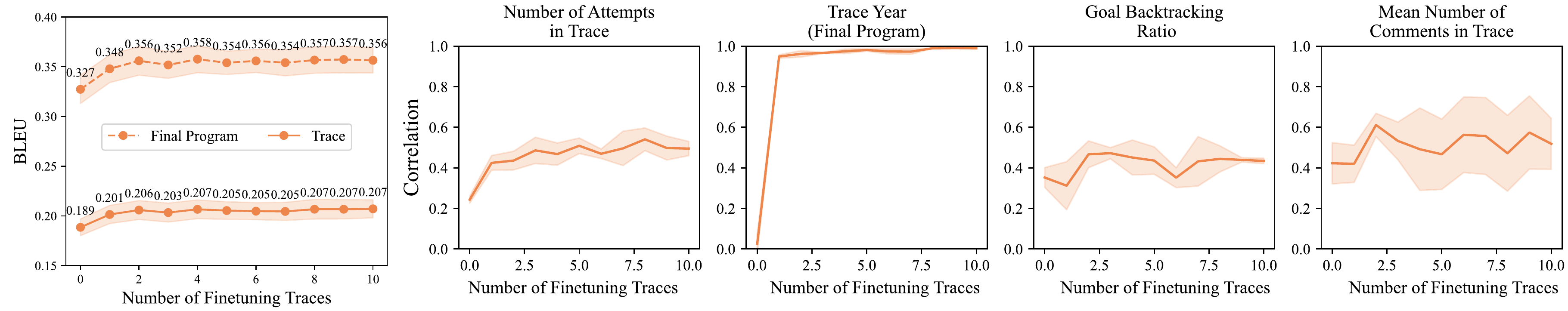}
    \caption{\textbf{Adaptation to New Students.} The leftmost plot shows BLEU score, and the right plots show correlations for different code properties. Results are for finetuning just student embeddings of the \all model. We report means and standard errors of the mean across 3 random seeds.}
    \label{fig:finetuning_results}
\end{figure}

\subsection{Adapting to New Students}
\label{s:finetuning}

Although our model cannot predict behavior for entirely new students from IDs alone, efficient few-shot personalization is essential for deployment in educational settings where new learners continuously join the platform. Recall that each new student ID is initially mapped to a random 768-dimensional vector via the student embedding layer. We evaluate how efficiently—both in terms of data and model parameters—the \all model adapts to new students.

We select all unseen users from the \unseenstudentseenprogram split with 20–200 traces and divide them into two groups: 5\% for hyperparameter selection and 95\% for finetuning and evaluation. For each student, we order traces chronologically and, for each $k \in [1,\dots,10]$, train on the first $k$ traces using early stopping, then evaluate on the remaining traces.
We hold all parameters fixed except the student embedding layer, updating only the weights of the student MLP module associated with the students whose traces are finetuned on.

\paragraph{Results} As shown in Figure \ref{fig:finetuning_results}, there is an increase in BLEU score for both final programs and full traces with just a single finetuning trace per student. We observe that BLEU scores stagnate after around k=4 finetuning traces. We also observe a sharp increase in correlation for trace year, followed by stagnation, suggesting that this property is learned quickly. For number of attempts in a trace, correlation increases, while improvements for mean number of comments and goal backtracking ratio are more mixed, with correlations increasing slightly until k=2, then stagnating. 
\subsection{Error Recovery and Model Control}
\label{s:error_recovery}

\begin{figure}[t]
    \centering
    \includegraphics[width=0.99\linewidth]{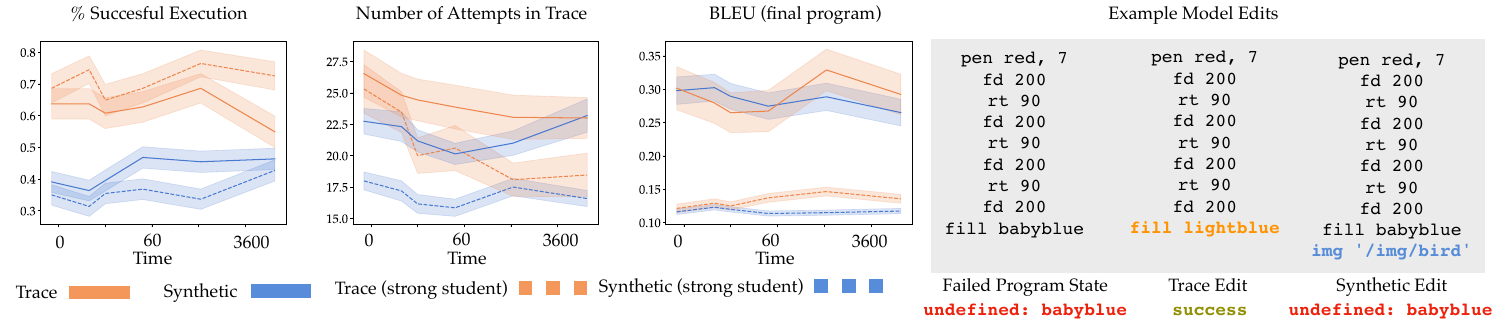}
    \caption{\textbf{Model Controllability and Error Recovery} We sample program traces from the \all and \synthetic models by conditionally generating on prefixes consisting of randomly sampled program states that failed to execute, held-out from training. We vary the amount of time between the broken trace and the next program, and compare the effect of student embedding (dashed vs. solid). Shaded area represents standard error of mean, and we show example edits from both models (``\texttt{babyblue}'' is an undefined color).}
    \label{fig:error}
\end{figure}

Finally, we evaluate the ability of the \all model to help students recover from errors and investigate whether these abilities can be controlled.
We sample mid-trace programs that fail to execute 
written by held-out students and create inputs with their student ID, the trace title, the sequence of program states up until that program, and the time header for the next program state (\eg \texttt{CODE 6:(2018-10-19 14:12:38)}. We then conditionally generate the remainder of the trace using either the \all or \synthetic model and measure: whether the final program state successfully executes, the number of attempts taken, and the BLEU score between the generated and ground truth final programs.

Additionally, we compare the effect of two kinds of model controllability on error recovery. First, we vary the time header in the prefix: If the failed program has the header \texttt{CODE 5: 2018-10-19 \textcolor{red}{14:12:37}}, we select a time $t \in [0.5, 1, 5, 60, 6400]$ in seconds which we add to the header (\eg \texttt{CODE 6: 2018-10-19 \textcolor{red}{14:12:38}} for $t=1$). Second, we vary replacing the ground truth student ID with a "strong student" embedding, selected by choosing the student in our training dataset with the highest number of program traces and lowest degree of goal backtracking. 

\paragraph{Results} In Figure \ref{fig:error}, we find that the \all model generates traces that reach a successful program state more than \textbf{60\%} of the time, outperforming the \synthetic model, which can only append lines as edits. Furthermore, replacing the student ID with the strong student embedding (dashed lines) {increases} the rate of successful execution of generated final programs, but only for the \all model, confirming that the student embedding carries useful information about strong student edit behavior. Interestingly, we observe the opposite when evaluating the BLEU score: using the strong student embedding decreases the ability for both \all and \synthetic models to reach final program states that are similar to the ground truth, showing that the learned student embeddings carry important information for personalization.
Finally, although increasing the amount of time between the failed program state and the rest of the trace does not appear to affect successful code execution, it leads to fewer numbers of attempts for the \all model. This indicates that we can control the granularity and extent of program modifications, which can be useful for adapting to different styles of feedback and interventions.  

We additionally trained a more complex \texttt{synthetic-complex} model where the synthetic edits can either add or delete any number of lines from any location in the current program state. Therefore, any two arbitrary program states can be connected with a sequence of edits. We find that, for error recovery, the \texttt{trace} model still outperforms the \texttt{synthetic-complex}  model, which only slightly outperforms the \texttt{synthetic}  model at lower values for time (x-axis, see Figure \ref{fig:error}). Results for  \texttt{synthetic-complex}  include the following values:
\begin{table}[h!]
\centering
\caption{Performance across different time scales.}
\begin{tabular}{c c c c}
\hline
Time (s) & Trace \% Correct & Synthetic \% Correct & Synthetic-Complex \% Correct \\
\hline
0.01 & 0.64 & 0.39 & 0.41 \\
1    & 0.60 & 0.39 & 0.40 \\
60   & 0.64 & 0.45 & 0.40 \\
\hline
\end{tabular}
\end{table}

We believe the reason for this poor performance is that the method for creating \texttt{synthetic-complex}  treats all lines equally, and is not able to capture how some aspects of a program (\eg function definition) are more susceptible to incorrect implementations than others, hindering error recovery.

\section{Related Work}
Motivated by concerns that AI tools such as large language models (LLMs) might foster student over-reliance by reducing student engagement \cite{bastani2024,nie2024gptsurprise}, there has been increasing interest in understanding how to effectively use LLMs within educational contexts, such as generating debugging hints in programming contexts \cite{kotalwar2024hintsinbrowser}. These works build on an extensive line of research that studies how computational modeling tools can successfully capture how student knowledge evolves over time \cite{piechknowledgetracing}, as well as identify different stages of student behaviors when learning programming, such as ``tinkering'' versus planning \cite{berland2013using}.

Closest to this work are papers that propose methods to learn ``edit embeddings'' from student code edits; however, they largely consider smaller datasets that do not contain student behavior across assignments as diverse as in \pc \cite{heickal2025learning}.
Furthermore, these works do not address whether features useful for personalized edit modeling can only be learned from real student edit behavior, versus the \synthetic or \last baselines we compared against. Other datasets capturing student programming behavior include BlueJ Blackbox (\cite{brown2018blackbox}), which captures extremely granular IDE-level actions; FalconCode (\cite{eliseeva2023falconcode}), which contains only final submissions rather than intermediate states; and StudentEval (\cite{khoja2024studenteval}), which captures student prompts. Finally, concurrent work by \cite{miroyan2025parastudent} evaluates how LLMs finetuned on a much smaller dataset (< 1M traces) generates traces aligned with real student traces along several properties (\eg error patterns). However, they do not go beyond surface-level alignment or disentangle which code properties can be learned as features of students versus program assignments.



Beyond education, our work connects to the literature on LLM reasoning \cite{cot}. These works seek to improve end-task performance by conditioning on intermediate reasoning steps, either by training models to reason \cite{star} or by sampling reasoning traces at inference time \cite{wang2023selfconsistency}. 

\section{Conclusion and Future Work}

We introduce a dataset of coding traces written by real students on \pc and show that models trained on full edit traces learn stronger representations of students’ coding behaviors than models trained on synthetically generated traces or final programs. This focus on modeling student behavior in an educational context departs from prior work on code generation, which has largely emphasized task accuracy.

Modeling individual student behavior from edit traces has important educational implications. Our results show that models can predict when students will backtrack or struggle with specific concepts (Section 4.2), enabling early interventions such as detecting whether an assignment is appropriately challenging. Efficient adaptation to new students with just 2–4 traces (Section 4.4) makes this feasible in real classrooms, where teachers cannot manually model each learner. Moreover, using learned student representations to recover from errors while preserving individual coding style (Section 4.5) addresses a key educational challenge: providing personalized feedback that respects students’ learning trajectories rather than enforcing a single “correct” approach. These findings open the door to intelligent tutoring systems, automated hint generation, and assessments that evaluate learning processes, not just final correctness.

\subsubsection{\ackname}

We thank \pc for their help with access to anonymized user logs.
This work was sponsored by Intel and the National Science Foundation under grants IIS-2212310 and CCF-2217064, and an MIT AI for Intelligence and Augmentation seed grant.  AR is supported by  NSF GRFP 2023357727. JA is additionally supported by a Sloan Research Fellowship. 




\bibliographystyle{splncs04}
\bibliography{splncs04}

\end{document}